# TOWARDS AN ETHICS-AUDIT BOT

Siani Pearson, Martin Lloyd
*Just Algorithms Action Group (JAAG), UK*
siani.pearson; info@jaag.org.uk

Vivek Nallur
*School of Computer Science, University College Dublin, Ireland*
vivek.nallur@ucd.ie

**ABSTRACT**

In this paper we focus on artificial intelligence (AI) for governance, not governance for AI, and on just one aspect of governance, namely ethics audit. Different kinds of ethical audit bots are possible, but who makes the choices and what are the implications? In this paper, we do not provide ethical/philosophical solutions, but rather focus on the technical aspects of what an AI-based solution for validating the ethical soundness of a target system would be like. We propose a system that is able to conduct *an ethical audit* of a target system, given certain socio-technical conditions. To be more specific, we propose the creation of a bot that is able to support organisations in ensuring that their software development lifecycles contain processes that meet certain ethical standards.

## 1. INTRODUCTION

There is a complex relationship between AI, ethics and audit. AI refers "to any technology that performs tasks that might be considered intelligent – while recognising that our beliefs about what counts as 'intelligent' may change over time" (Nuffield, 2019). Recently, AI has come to refer just to machine learning (ML), but it can also encompass other techniques, including rules-based reasoning. It brings new opportunities but can also heighten ethical problems (IEEE, 2019; Munoko, 2020). Most ethical guidelines, principles for AI etc. seem to be focused on people building AI systems. In contrast, building AI systems that are themselves ethical has not received as much attention. There have been some attempts at adding ethical reasoning (rule-based, constraint-satisfaction, etc.) (Nallur, 2020), but not enough at testing systems for possible ethical violations. The main contribution of the paper addresses this issue by means of the proposal of an 'audit bot' solution. In the following sections we consider current usage of AI for audit, and then the need to define what ethics audit requires from AI to function properly, followed by the proposal of the ethics Audit Bot that we are designing and concluding remarks.

## 2. AI FOR AUDIT

AI can potentially increase efficiency of audit, provide greater evidence and insight into operational processes and create competitive advantages. Many of the larger auditing companies are already using AI and it already has a vital role in forensic accounting (for example, the 2016-20 investigation into corrupt payments made by some Airbus staff[1]). Condition monitoring, statistical analysis and business intelligence are embodied in a variety of computer-assisted audit tools and techniques that form a growing field within the IT audit profession. The term 'audit bot' is already being used within the audit realm[2], but typically just to validate system data, determine conclusions through logical checks, address problems such as poor password management, generate documents and information requests, and input received data into systems.

In the security domain, security information and event management provide real-time analysis of security alerts, log security data and generate compliance reports. It is standard practice to integrate abnormality detectors in security tools, using ML techniques, but due to the use of these techniques the system will tend to replicate bias and historical undesirable features.

## 3. ETHICS AUDIT

Organisations undergo or use many different types of audit. An *ethics audit* is "the systematic, independent, and objective examination and evaluation of the ethical content of the object of the audit" (Kaptein, 2020). Thus, it is carried out in a way that provides essential conditions for the users of an ethics audit to trust the conclusions of the audit, and this includes a trusted entity (usually a suitably qualified professional) conducting the audit; by implication, all tools used in the audit shall be trustworthy as well.

Ethics audit is designed to dig deep into the practices of a business, to see how closely an organisation follows its rules, and conforms to or exceeds the ethical standards of its industry and society. An ethics audit (with clear definition of its scope and a committee established to oversee it) is increasingly seen as part of corporate social responsibility and governance; complementary aspects include incorporation of ethical values into organisational policies, high level support from top managers and the board of directors, and more proactive measures such as ongoing ethical impact assessment and design across the organisation's activities, and good security. Limited information exists about ethical audits carried out in practice (Kaptein, 2020).

### 3.1 Different Types of Ethics Audit

An ethics audit is necessarily more qualitative and subjective in nature than compliance checking, which considers adherence to legal requirements in given contexts. Auditing against high level, general principles such as those espoused in ethical guidelines (Hagendorff, 2020) is difficult to do, and especially if this activity involves trying to drill down into low level events occurring within the system. Therefore, we recommend "domain specific ethics", such as for bioethics (Beauchamp and Childress, 1991). Not only are these typically tailored to the social milieu but also easier to map to lower-level activities/events in the domain. Different approaches using technology in auditing ethical aspects are possible, including:

- Providing a tool that assists users in thinking about ethical issues and capturing their responses in a secure way that can be reviewed later. For example, based on an "Ethics Canvas" (ODI, 2020).
- Mapping high level ethical values to lower-level incidents that can be checked by software: this is extremely challenging and often of limited value (Pearson and Allison, 2009).
- Conversion of legal or other requirements into modal logic and using automated reasoning, with forms of conflict resolution, to contextually recommend actions. In practice, though, these have proven difficult to populate with encoded rules, keep up to date as legislation changes, and train and motivate users.
- Encoding organisational policies that already embed ethical values and legal requirements appropriate to the organisation, and using these policies within semi-automated enforcement and compliance checking.
- Multiple software agents, each with their own goals and values, perhaps representing different persons.
- Using a ML system trained to recognise anomalies to detect high risk activity and flag this up at an appropriate time for human users to consider.
- Broadly speaking, permitted exceptions audit may be achieved via anomaly detection techniques, obligations audit via model checking, and access legitimacy audit via query based reasoning and pattern matching (Reuben, 2016).

---

[1] https://www.bbc.co.uk/news/business-55306139
[2] https://auditbots.com/about-us/

## 3.2 Ethical Audit Bots

A bot is an application that performs an automated task. Correspondingly, different kinds of *ethical audit bots* are possible, including one or more of the following potentially acting together:

(a) audit bot(s) running in the background within an organisation's systems, producing a secured repository of evidence that supports a human ethical auditor, highlighting events that might be considered unethical by monitoring (non)compliance with rules in which ethical values have been encoded

(b) audit bots that focus on detecting breaking of the rules and considering whether or not that is justified (for example, in an emergency or because of safety)

(c) a recommender bot, flagging up anything likely to have ethical repercussions and/or future recommendations about change of behaviour to a human based on analysis of logs: for example, it might detect that offensive language was being regularly used by an employee. This itself may cause privacy concerns and hence it should be clear to all employees what data is accessible to the bot.

The object of an ethics audit can differ, according to the judgment of the entity setting the scope. It may be set in terms of intentions, behaviour or consequences, following the main distinction of ethical theories in terms of virtue ethics, deontology and utilitarianism respectively (Harris, 2008). Choices have to be made, but by whom (DuQuenoy, 2005)? The answer is contextual, but in general it is better to have a more flexible and transparent system where: individual stakeholders can reflect their ethical values into the system; ethical encoding is robust and understandable to other parties; stakeholders or their representatives are involved within the ethical decision-making process, with potential harm flagged up to users to consider (IEEE, 2019).

## 4. EXAMPLE OF AI FOR ETHICAL AUDIT: 'AUDIT BOT'

We propose the introduction of Ethical Audit Bots to perform automated audit supporting analysis of the extent to which protection and social justice is being achieved during the design and/or operational support phases.

During development, one or more Audit Bots would be introduced into the processes of specifying, programming, verification, validation, and release of decision-making systems. Audit Bots will be used throughout the software life cycle and therefore applied during the operational support phase when the system is live and requires maintenance and or modification. An ethical Audit Bot needs to be assessed at all stages of its life cycle – just like the systems it is intended to monitor; for example, any proposed monitoring of employees by the audit bot needs to be assessed to ensure that this would be ethical.

This high-level vision may be addressed in different ways. The Ethical Audit Bot may gather and aggregate evidence in a form that is useful for a human auditor, potentially as part of an audit process mapping to new protection standards. Or it may, for example, be able to flag up harms itself, with a justification for the final outputs. With regard to risk, there are already several approaches that could be used. For example, ALARP (as low as reasonably practicable) categorises risks into intolerable; so small as to be acceptable; and those which lie in between, which must be reduced to the lowest practicable level (UK HSE, 2020). Mitigating the risks in the ALARP category concerns the balance between the cost of the risk and the cost of mitigation. This brings in the deep issues of the cost of harm. Some of the audit is likely to be very context specific; analogously in the field of sustainability accounting standards, new reporting metrics are currently being developed by the World Economic Forum (Tysiac, 2020).

What rules do we need to achieve ethical operations at all stages of the lifecycle? During the drafting of the system development part of the system lifecycle, both the questions to be investigated and the range of expected answers need to be defined, as part of a safety, ethics and quality plan. Once defined, at the next part of the system lifecycle an Audit Bot is built to automate this, in the sense of answering the questions for itself and building a log of the answers that can be shown to a human. These can be designed not only to drill down into the details, but also helping to flag up where there are particular concerns. It might be designed to flag up high harm events immediately and/or to produce reports periodically and/or on demand. An automatic auditor could run in the background.

If an organisation publicly commits to these rules, then it should also specify how the internal processes of the organisation will change to accommodate this. Once this is done, job descriptions will change and the bot can monitor whether the employee performed the activity that was expected of them.

Here are some diverse examples of what the Audit Bot could be doing:
- (*pre-operational*): Checking whether any data set used for training a ML system was assessed for bias.
- Checking that at least one person on each project team must have ethics training and there is a recorded process about resolving ethical issues; integration with an ethical issue tracking system, whereby ethical issues can be flagged up, a record kept, and appropriate responsibilities clearly assigned. Tracking this over time, ascertaining whether a software build happened before these issues were addressed.
- Auditing the due diligence that actors perform in order to make more informed decisions about ethical actions as part of their role, e.g., ML practitioners selecting from standardised datasheets listing the properties of different training data sets (Gebru et al, 2018).
- (*operational*): Every time a document is reviewed and checked into the document management system, checking that the review's author appears in the competence register that every entity participating in the project has to keep up-to-date (as for the safety domain), and so is qualified to perform that review; also checking in an independence register to determine whether the reviewer is sufficiently independent for this level of severity or safety of this system.
- Monitoring employees' online activity to detect excessive working hours.
- Preventing gender biased job descriptions going out.
- Using measures that check whether systems have changed significantly since a previous point, e.g., when they were initially trained; based on this, recommending if there is cause for alarm and the ML model might need to be changed.

This illustrates the various different types of checks that could be made. This approach allows greater accuracy, with the bot building the evidence of good practice in real time in a sealed repository which can then form the basis of building a case usable in court, to help justify why, for example, the systems used to build systems are ethical. It is intended to handle the lower-level routine tasks of pattern matching and recording, rather like the way static analysis of code picks up the more easily detected errors prior to review, although advances in technology have shown that some specialised analysers such as KLEE (Cadar et el, 2008) outperform humans in their bug detection capability.

## 5. CONCLUSION

We are working to develop an Audit Bot as a novel mechanism for ethical audit, in a way that takes tedious work away from humans, as well as being able to drill down and provide levels of evidence that might not be readily obtainable by workers. The authors' experience of auditing in the safety domain suggests the proposed approach would have benefit in the ethical domain by getting humans to think about their answers and actions, providing integrated and customised help, logging evidence about decisions made, and storing evidence securely.

If an organisation publicly commits to some principles (such as reducing bias to a low level), then it must also be possible to ask what processes it has put in place to ensure that these principles will be upheld. Using the proposed approach, once processes are defined, they can be traced back into job descriptions all the way down the employee chain. So, the Ethical Audit Bot should be able to pick up from employee-activity logs, where in the chain of command, the process failed (or was not followed); this helps real people to be held accountable in the case of an audit failure, and effort directed to avoid similar problems arising in future.

The Ethical Audit Bot is targeted at testing *any* kind of system (not necessarily limited to those employing AI techniques) for possible ethical violations. As with development of many other complex AI techniques, steps towards full automation are being done in an incremental way. In the short term, the parts that can be automated by the Ethical Audit Bot are much narrower in scope than a human-based ethics audit; these should therefore be considered supplementary to a human auditor. Automated checking to increase evidence is closely allied to checking the organisation's behaviour, corresponding to a deontological ethical approach, but other ethical perspectives apply too, notably the virtue ethics approach, whereby individuals' education is a key factor.